\documentclass[conference]{IEEEtran}
\IEEEoverridecommandlockouts
\usepackage{cite}
\usepackage{amsmath,amssymb,amsfonts}
\usepackage{algorithmic}
\usepackage{graphicx}
\usepackage{textcomp}
\usepackage{xcolor}
\usepackage{subfigure}
\usepackage{balance}
\def\BibTeX{{\rm B\kern-.05em{\sc i\kern-.025em b}\kern-.08em
    T\kern-.1667em\lower.7ex\hbox{E}\kern-.125emX}}
\begin{document}

\title{Quantified Facial Temporal-Expressiveness Dynamics for Affect Analysis\\
}

\author{\IEEEauthorblockN{Md Taufeeq Uddin}
\IEEEauthorblockA{Department of Computer Science and Engineering\\
University of South Florida\\
Tampa, FL 33647, USA\\
Email: \small \texttt{mdtaufeeq@mail.usf.edu}}
\and
\IEEEauthorblockN{Shaun Canavan}
\IEEEauthorblockA{Department of Computer Science and Engineering\\
University of South Florida\\
Tampa, FL 33647, USA\\
Email: \small \texttt{scanavan@usf.edu}}}


\maketitle

\begin{abstract} 
The quantification of visual affect data (e.g. face images) is essential to build and monitor automated affect modeling systems efficiently. Considering this, this work proposes quantified facial Temporal-expressiveness Dynamics (TED) to quantify the expressiveness of human faces. The proposed algorithm leverages multimodal facial features by incorporating static and dynamic information to enable accurate measurements of facial expressiveness. We show that TED can be used for high-level tasks such as summarization of unstructured visual data, and expectation from and interpretation of automated affect recognition models. To evaluate the positive impact of using TED, a case study was conducted on spontaneous pain using the UNBC-McMaster spontaneous shoulder pain dataset. Experimental results show the efficacy of using TED for quantified affect analysis.

\end{abstract}

\section{Introduction}
\label{intro}

Visual expressiveness is the representation of how people outwardly display how they feel (e.g. visual emotion elicitation), which has some degree of variability based on the affective response \cite{kring2007facial}. Complementary to this, facial temporal dynamics is how people change their facial expressiveness to display the transition of emotional states such as negative state to neutral state to positive state. Considering this, we combine static and dynamic facial information to quantify the temporal-expressiveness of faces for robust affect analysis.  

There are multiple challenges in automated emotionally intelligent systems that necessitate the quantification of expressiveness. First, affect datasets are usually labeled by human reporters using observer-report, self-report, or expert-report. Discrete categories for a segment of data (e.g., happy label for video) are used for these, which is a subjective task. The major concern is that this approach fails to capture details of affect information from the samples since the intensity of expression may not be the same throughout the video \cite{gunes2011emotion}. To understand and build better automated affect modeling systems, we should measure the intensity of the facial-emotion expression along with the discrete expression. Second, to develop and monitor intelligent systems, practitioners have to perform tasks such as data collection, data processing, modeling, inference, and system monitoring. Due to the large amount of unstructured data (e.g., video) that is currently available \cite{zhang2016multimodal, cheng20184dfab}, it is infeasible for practitioners to manually explore the data to investigate the quality of data (e.g. expressiveness). Therefore, there is a need for automated methods that help practitioners explore the dataset efficiently. Also, having measurements such as facial expressiveness, that can create human-understandable interpretations of the decisions from automated affect models is essential, especially in sensitive domains like healthcare.

The quantification (i.e., objective measurement) of expressiveness is important from the application domains perspective. Even though subjective measurements such as self-report (considered the gold standard in psychology), and observer report (controlled and goal-oriented response) has been quite extensively used in practice, these reports could be the victim of reporters bias, variance, and ambiguity based on memory and verbal ability \cite{walter2013biovid, walter2014automatic, williams2000simple, uddin2020multimodal, williamson2005pain}. Also, unconscious people, people with mental disabilities, and children may not be able to provide subjective reports. Healthcare professionals need objective pain assessment methods to treat patients objectively. The objective, quantified measurement of expressiveness has the potential to assist people with communication disabilities \cite{macari2018emotional}, people with emotional expression deficits \cite{wang2008automated}, and people with sleep deprivation issues \cite{minkel2011emotional}. Last but not least, it has the potential to improve peer competence \cite{lindsey2019frequency}, anticipation and interaction \cite{liu2019role} among people.

\begin{figure*}
\setlength{\belowcaptionskip}{-2ex}
\includegraphics[width=.95\linewidth]{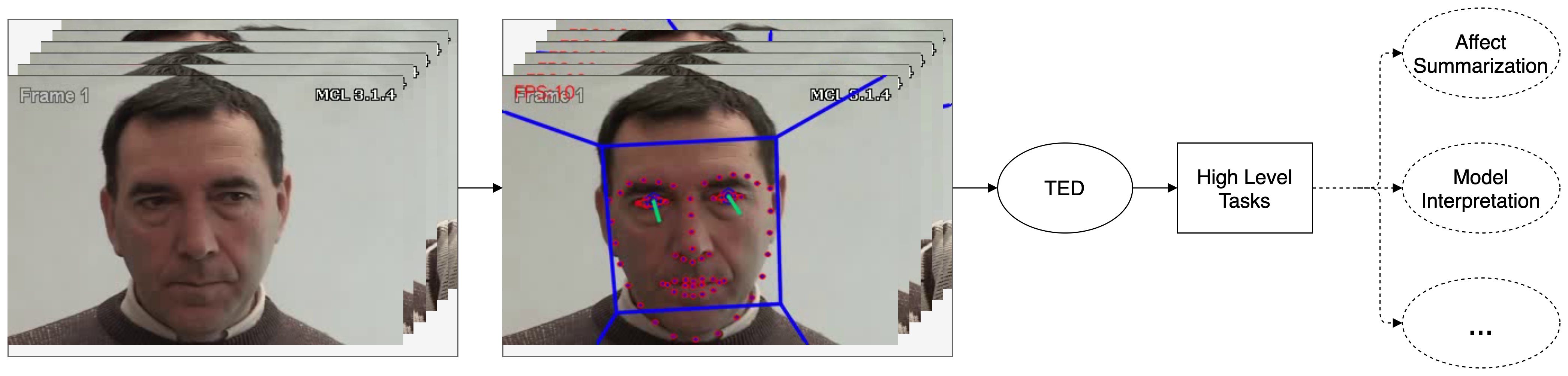}
\label{methodDiagram}
\caption{Computational pipeline of TED. From the input video, multimodal facial features (e.g., landmarks) are computed and then used as input to TED. The output is used to perform high-level tasks that include, but are not limited to, affect summarization, and model interpretation. NOTE: (...) indicates other tasks that TED can perform such as domain shift identification.}
\label{fig:pipeline}
\vspace{-5mm}
\end{figure*}

In this work, we propose quantified facial Temporal-Expressiveness Dynamics (TED) to quantify facial expressiveness at the video frame level by incorporating static and dynamic facial information. This is done by leveraging multimodel features including facial action unit \cite{ekman1978Facs} intensities, facial landmarks, head pose, and eye gaze. The features are used as input to TED, which then outputs a relative, continuous score. This score can be computed for the overall expressiveness of a frame, and/or affect-specific expressiveness for a given frame. Based on the computed expressiveness score, high-level tasks can be completed such as i) creating a summary of affect dataset; ii) setting up the expectation from affect prediction models and interpreting the results of the model. To evaluate the efficacy of TED, a case study was performed on spontaneous pain using the UNBC-McMaster shoulder pain dataset \cite{lucey2011painful} and obtained promising results. This work has the potential to complement and/or to replace existing measurement methods \cite{walter2013biovid, walter2014automatic, williams2000simple, williamson2005pain}, to identify the moment of interests \cite{Lei2020} in videos, to analyze affective visual data, and to interpret affective machine learning models.

Our main contributions can be summarized as follows.

\begin{enumerate}[1.]
    \item Quantified facial Temporal-expressiveness Dynamics (TED) is proposed to quantify facial temporal-express\-iveness. See Fig. \ref{fig:pipeline} for the computational pipeline of TED. Quantitative and qualitative analyses demonstrate the efficacy of TED.
    
    \item Summarizing unstructured visual data using TED, to extract insightful information from the affect data, is introduced. Useful insight such as visual cues may not always be a strong indicator of affect state are captured.
    
    \item Expressiveness-based interpretation for predictive affect models is introduced. The results suggest that TED can be used to interpret the goodness of predictive models.
\end{enumerate}

The rest of the paper is organized as follows: Section \ref{relatedWork} covers the context and background information. Section \ref{TED} contains the proposed algorithm and its applications. Section \ref{dataset} contains a brief description of the studied dataset, and Section \ref{exprNRes} comprises the conducted experiments and obtained results. Finally, in Section \ref{DisLiF}, the generalization of the proposal is described and the limitations are addressed.

\section{Related Work}
\label{relatedWork}

\textbf{Expressiveness and temporal dynamics}. Although less extensive, there are some important works on facial dynamics in which dynamics was defined for individual action units using discrete categories such as onset, apex, offset. Jiang et al. \cite{jiang2013dynamic} and Valster et al. \cite{valstar2007combined} both predicted AU temporal segments using temporal appearance features. There are some works on expressiveness as well. Such as Hammal et al. \cite{hammal2018facial, hammal2019dynamics} examined facial expressiveness in infants with and without complex congenital conditions named craniofacial microsomia which could impair facial expressiveness; they reported that expressiveness varied between positive and negative affect. Neubauer et al. \cite{neubauer2017manual} reported that oxytocin increases facial expressivity in both healthy and schizophrenia populations. Guha et al. \cite{guha2016computational} investigated how children with high functioning autism (HFA) differ from their typically developed peers; they reported a reduction in the complexity of facial behavior especially in the eye region of the HFA group compared to their peers. Lin et al. \cite{lin2019context} proposed an extension to the BP4D+ dataset \cite{zhang2016multimodal} in which they rated video sequences in terms of expressiveness using a discrete scale and rely on the human rating from crowd workers at Amazon Mechanical Turk. They studied only segments of video where emotion elicitation was at its peak. Werner et al. \cite{werner2017analysis} provided some insight into expressiveness when the pain was induced via heat. One of their interesting findings was that low pain intensities do not show facial response in many cases. Lei et al. \cite{Lei2020} reported that expressivity can explain unexpectedness of outcomes in social dilemmas and also can identify the moment of interest in video sequences. In contrast to these studies, we propose to characterize expressiveness more objectively to incorporate as much information as possible using an algorithmic approach to produce a continuous temporal-expressiveness score at frame level for a given affect video. There are several works on expressiveness in affective robotics (e.g., expressive humanoid robot with facial muscle movements \cite{faraj2020facially} to improve qualitative interactions) and affective speech processing (e.g., expressive speech synthesis from text \cite{dahmani2019conditional}), as well.

\textbf{Annotated visual affect datasets}. The publicly available affective computing datasets usually contain either categorical labels (e.g., happy, sad), dimensional labels (e.g., arousal, valence), or both. Data are usually annotated by participants, observers, and domain experts. For instance, BP4D+ is a spontaneous facial expression dataset that was annotated at the video sequence level by both observers and participants. BP4D \cite{zhang2014bp4d} is also a spontaneous facial expression dataset and annotated by observers. There are several other static and dynamic facial expression datasets that include, but are not limited to, FERA \cite{valstar2015fera} (static expression), BU3DFE \cite{tang20083d} (static expression), and BU4DFE \cite{zhang2013high} (dynamic expression). Some specialized datasets are collected to study facial action units (muscle movements) \cite{ekman1978Facs}. For example, DISFA \cite{mavadati2013disfa} contains spontaneous facial expressions and manual annotations of a subset of action units using a $6$ point scale in the range of $[0, 5]$. DISFA+ \cite{mavadati2016extended} is an extension of the DISFA dataset that contains posed and non-posed expressions along with all other features of DISFA. FEAFA \cite{yan2019feafa} is a recently released posed expression dataset that contains expert-coded action unit intensities in the range of $[-1, 1]$.

There are also datasets that were designed to study facial pain expressions such as MIntPain \cite{haque2018deep}, BioVid \cite{walter2013biovid}, emoPain \cite{aung2015automatic}, and UNBC-McMaster shoulder pain expression datasets \cite{lucey2011painful}. UNBC-McMaster shoulder pain expression dataset \cite{lucey2011painful} is a unimodal dataset that contains the spontaneous facial expression of patients' with shoulder pain. \textit{In this work, we experimented with UNBC-McMaster pain dataset} given the dataset has adult patients' (as participants) elicitating spontaneous facial expressions, manually annotated intensities of (pain-related) facial action units of each frame, pain intensity annotation using pain assessment scale, self-reported and observer reported pain scores, and also data collection was done in a laboratory setting that was similar to hospital.

\section{Method} 
\label{TED}

\subsection{Motivation and Multimodal Facial Features Tracking}

The first step of the proposed method is to track multimodal facial features. Constrained local neural fields (CLNF) \cite{baltrusaitis2013constrained} is used to track 2D facial landmarks. It is known that people tend to move their heads and faces when they make a transition from one affective state to others (e.g., neutral to positive affect state) \cite{hammal2018facial, hammal2019dynamics} and this information can be captured from head pose \cite{bassili1979emotion, knight1997role}. Considering that, this work tracks head pose translation and rotation using CLNF \cite{baltruvsaitis2016openface}. 

Attentiveness (gaze) is also a crucial part of affect analysis, especially for people with communication disabilities \cite{macari2018emotional}. Recent works showed a strong link between emotion recognition and eye gaze estimation \cite{awad2019role, wu2019continuous} which motivates us to use gaze to measure temporal expressiveness dynamics. Thus, we incorporate eye-gaze features in our algorithm by capturing gaze from the left eye and right eye. 

Facial action unit (AU) \cite{kring2007facial} intensities can represent both static and dynamic information about an expression. Motivated by this, a core component of TED is AU intensities. Note that AUs have a strong relationship with individual expressiveness, personality, skin conductance, heart rate, and self-report \cite{ekman1978Facs}. Considering this, we analyze both manually coded AUs and predictive (machine learning) model coded AUs using the method proposed by Baltrusaitis et al. \cite{baltruvsaitis2015cross}.

\subsection{Quantified Facial Temporal-Expressiveness Dynamics}

In the proposed quantified facial Temporal-Expressiveness Dynamics (TED) algorithm, we combine static and dynamic facial information that is representative of the expressiveness since people use both static and dynamic information to convey emotion \cite{hadid2011analyzing}. To compute static information, this work relies on the facial action unit intensities \cite{kring2007facial, barrett2019emotional} for a given video frame.  Hence, in this work, for a given video frame, the static information is computed using Eqn. \ref{static}. 
\begin{equation}
    S = \sum_{1}^{n} e ^ v
    \label{static}
\end{equation}

Where $v$ contains AU intensities and $n$ is the total number of AUs. The argument behind computing the exponent of the intensities is to put more weight on AUs with high intensity such as extreme ($D$) and maximum ($E$) compared to low intensities such as trace ($A$) and slight ($B$). The motivation behind weighting is that, in terms of expressiveness, the difference between neutral (i.e. AU is inactive) and low intensities  (e.g. $A, B$) is relatively small, while the difference between low intensities and high intensities is relatively large. We encode the FACS AUs as follows:  $A = 1, B = 2,$ marked / pronounced $C = 3, D = 4, E = 5$. This is done to transform from a categorical scale to a numerical scale that TED can utilize. Our work is motivated by Werner et al. \cite{werner2017analysis}, who found that it is extremely hard to distinguish between neutral frame and low-intensity pain, while it is comparatively easier to distinguish between neutral and high-intensity pain. This is due to humans usually showing little to no facial response during low-intensity pain elicitation.

Note that dynamic information can capture coherency in affect identification, and can distinguish between posed and spontaneous expressions \cite{krumhuber2013effects}. To measure dynamic information, this work relies on the relative change of the proposed features (e.g., landmarks, headpose, gaze, and AU intensities). Since the theoretical grounding and mathematical representation (e.g. geometry, range of values) of landmarks, headpose, gaze, and AU intensities are different, the relative change is measured using vectorized standardized Euclidean distance, which is a unit normalized (i.e. independent) measurement technique \cite{tan2018introduction}. The relative change ($C_r$) between consecutive frames is computed using Eqn. \ref{dist}.
\begin{equation}
  C_r = 
  \begin{cases}
    0 & \text{if $var(f_i) + var(f_{i+1}) = 0$} \\
    \frac{var(f_{i+1}, f_{i})}{var(f_i) + var(f_{i+1})} & \text{otherwise}
  \end{cases}
  \label{dist}
\end{equation}

Where $f$ is the video frame, $i$ is the frame index, and $var$ is the variance \cite{diez2012openintro} between $f_i$ and $f_{i+1}$. Note that initial video frame $f_1$ is treated as a reference frame. Eqn. \ref{dist} does not provide information about the direction of change ($D_s$), therefore to get this information the sum of the vectorized displacement is computed using Eqn. \ref{sign} in which $m$ represents the length of the facial feature vector. 
\begin{equation}
  D_s = 
  \begin{cases}
    +1 & \text{if $\sum_{i=1}^{m} [f_{i+1} - f_{i}] \geq 0$} \\
    -1 & \text{otherwise}
  \end{cases}
  \label{sign}
\end{equation}

To accurately capture the dynamics, Eqn. \ref{dist} and Eqn. \ref{sign} are multiplied to obtain product $P = D_s * C_r$. After getting $P$ for the frames of the video, we then compute the moving average $M = \frac{1}{w} \sum_{i}^{w+i-1} P_i$ ($i$ is the video frame index) over $P$ using a window length of $w$. For example, when $i= 10$ and $w=5$, $M$ is computed over $P$ of video frames $10$ through $14$. It is important to note that $C_r$, $D_s$, $P$, and $M$ are computed for each facial feature set (e.g., landmarks ($L$), headpose orientation ($H_o$) and rotation ($H_r$), left and right eye gaze ($G_l$ and $G_r$), and AUs' intensities ($I$)), \textit{separately}.

Notice that static information $S$ is captured in Eqn. \ref{static} and dynamic information is captured in the form of moving average $M$ for each feature set. The static information $S_i$, in the $i^{th}$ frame, is combined with dynamic information $M$ computed over the frames up to $i$ for each feature set using a simple and efficient formulation (Eqn. \ref{tedScr}).
\begin{equation}
    Score = S_i * [1 + M_L * M_{H_o} * M_{H_r} * M_{G_l} * M_{G_r} * M_{I}]
    \label{tedScr}
\end{equation}

\begin{figure}
\centering     
\subfigure[Subject 43, sequence 05]{\label{fig:a}\includegraphics[width=.49\linewidth]{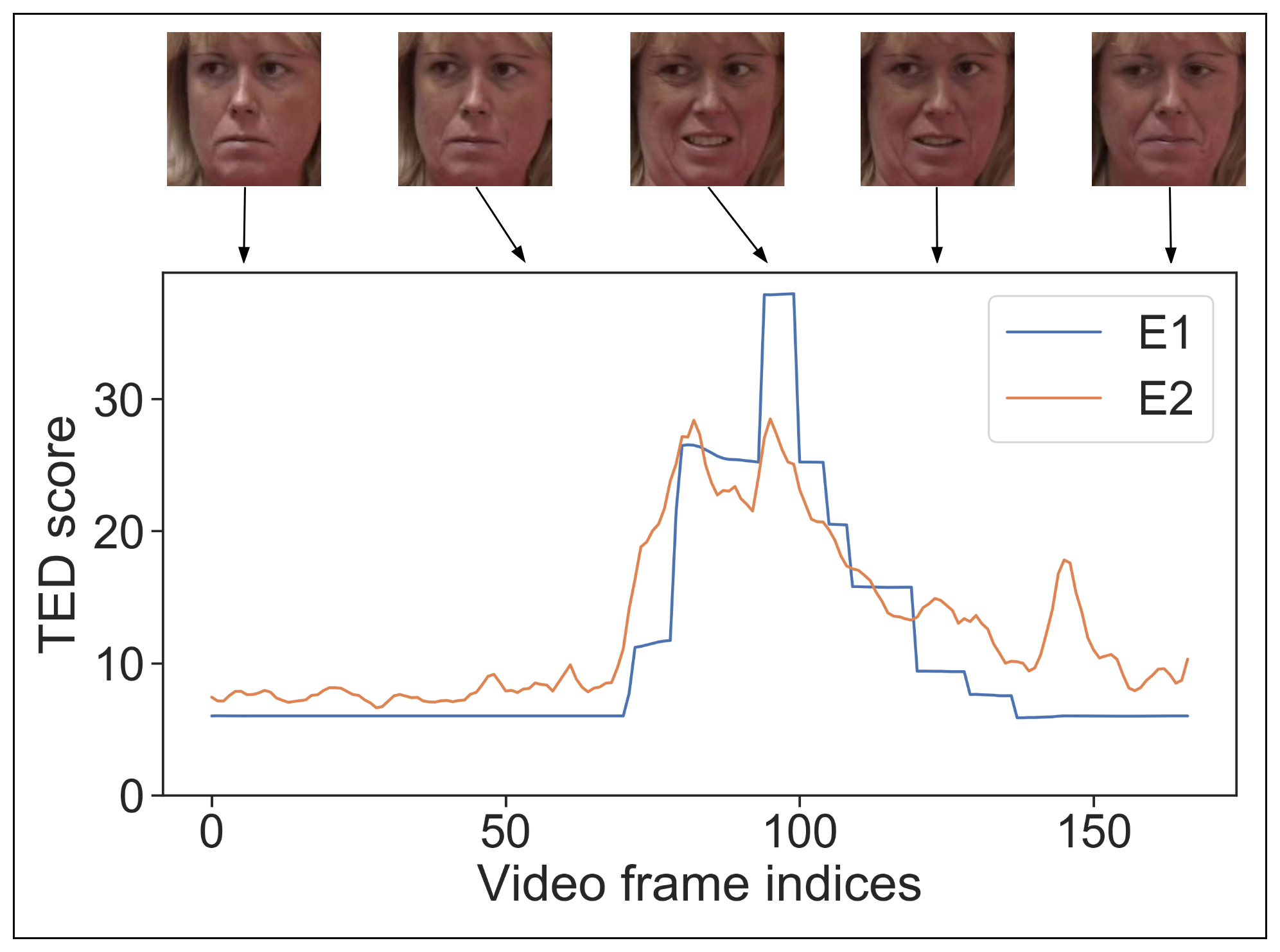}}
\subfigure[Subject 80, sequence 02.]{\label{fig:b}\includegraphics[width=.49\linewidth]{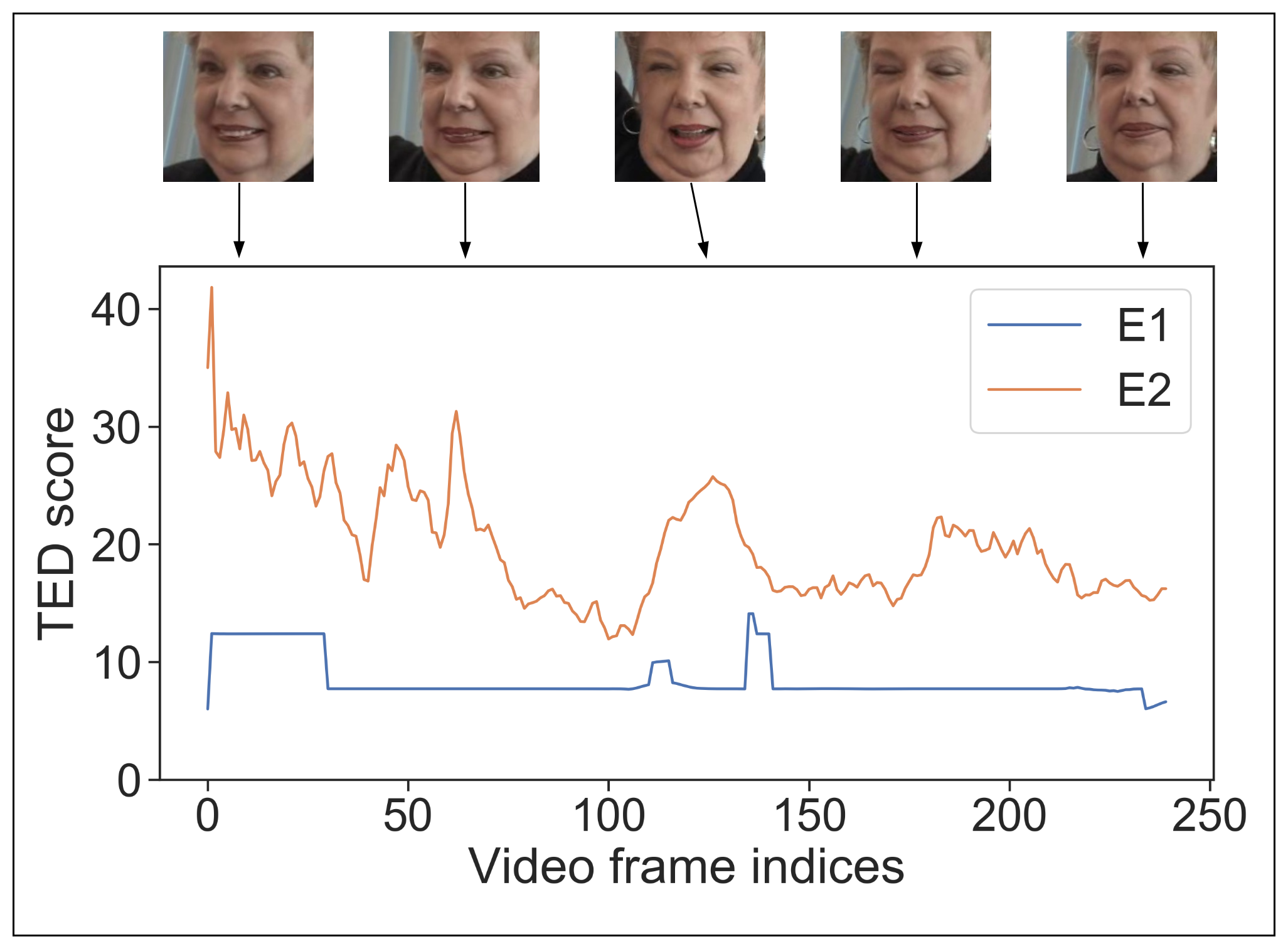}}
\caption{Sample TED scores computed in experiments E1 and E2. Difference between E1 and E2 is mentioned in Section \ref{exprNRes}.}
\label{sampleTEDscrCW}
\vspace{-5mm}
\end{figure}

Where $M_x$ represents relative change (dynamics) with respect to feature set $x$ (e.g. $M_L$ means the dynamics captured using the landmarks $L$). The output of Eqn. \ref{tedScr} is the quantified expressiveness score (TED score) of the frame $f_i$. See Fig. \ref{sampleTEDscrCW} for sample sequences with computed TED scores. 

\textbf{Overall facial expressiveness, and affect-specific expressiveness}. TED can be used to quantify expressiveness for a given human face in a video regardless of any specific affective expression, and in this work, we define it as "overall expressiveness". TED can also be used to quantify the expressiveness of affect-specific expression such as happy or pain. To compute affect-specific expressiveness, TED requires AUs that are related to that specific affective expression. For example, to compute the expressiveness of happy expression, TED requires AUs: $6, 7, 12, 25, 26$ as input. This is motivated by the work from Barrett et al. \cite{barrett2019emotional} which provided the list of combinations of AUs that correspond to affect-specific expressions.

\subsection{ High-Level Tasks (Use Cases of TED)}

\subsubsection{Affect Summarization} 
Having insight about data that are used, in affect modeling, is essential since the quality of the predictive models depends on the quality of data. Recall that visual affect data are unstructured data which makes the assessment of data quality extremely challenging. To alleviate this, TED can be used objectively at scale, within a very short amount of time, for data assessment as it maps video (unstructured three-dimensional data) to sequence of continuous values (structured one-dimensional data). Hence, this work outlines the summarization (profiles) of the visual affect data by performing statistical measurements on the expressiveness measured by TED. The proposed approach can create a descriptive summary of a video sequence, segments of a sequence, and the entire dataset. This has the potential to bolster the development cycle of automated affect systems by analyzing affect data at scale.

\subsubsection{Expectation and Model Interpretation}
Affective machine learning systems are expected and required to be interpretable to serve in sensitive application domains like healthcare. The system must have components that could assist professionals both from automated system development (e.g., software developer) and application domains (e.g., nurse) to understand how the model makes decisions (i.e., explain/interpret results). Since there is a strong relationship between expressivity and affect states \cite{werner2017analysis}, we devise the TED score to set up the expectation from the model, and to interpret the confidence of the model for a given sample. 
\begin{equation}
  E = 
  \begin{cases}
    \text{$Confidence$ ($\uparrow$)} & \text{when $Score$ ($\uparrow$)} \\
    \text{$Confidence$ ($\downarrow$)} & \text{when $Score$ ($\downarrow$)}
  \end{cases}
  \label{interpretScheme}
\end{equation}

For instance, in the case of pain localization from visual data, it is expected that when pain-specific TED score is high, the model should produce pain as output with high confidence (i.e., high pain class probability). In contrast, when TED score is low, the model is expected to generate low confidence for pain (i.e., high confidence for no pain) because a low TED score indicates less expressiveness, which means the sample is likely to be a no-pain sample (see Eqn. \ref{interpretScheme}). If the expectation is met, then we can say that there is an agreement between TED and the model (i.e., results likely to be correct). Otherwise, we can say that there is a disagreement between TED and the model, which means either the TED score or model prediction is likely to be wrong. That will lead us to further investigate TED and the prediction model, which will improve the trustworthiness of the system.
\section{Dataset}
\label{dataset}
The proposed TED was evaluated on the UNBC-McMaster shoulder pain expression dataset \cite{lucey2011painful}. The dataset contains $200$ video sequences ($48398$ frames in total) of $25$ patients elicitating spontaneous facial expressions. Each video frame in this dataset was manually AU coded using FACS via certified coders. The available AUs are $4, 6, 7, 9, 10, 12, 20, 25, 26, 43$. The dataset contains three types of self-reported pain scores at video sequence level using scales named visual analog scale (VAS), sensory scale (SEN), and affective (AFF). The dataset also has observer reported pain scores at video sequence level using observer pain intensity (OPI) scale. Note that these scales provide subjective measurement. Using VAS scale, patients label video sequence for emotional pain in the range of "no pain" (pain score $(P_s) = 0$) to "pain as bad as it could be" ($P_s = 10$). In SEN scale, patients reported pain from "extremely weak" ($P_s = 0$) to "extremely pain" ($P_s = 10$). In AFF scale, patients reported pain from "bearable" ($P_s = 0$) to "excruciating" ($P_s = 10$). In OPI scale (anchored Likert scale), trained observer reported observed pain expression from "no pain" ($P_s = 0$) to "strong pain" ($P_s = 5$). Facial pain intensity is also annotated at video frame level using the Prkachin and Solomon pain intensity (PSPI) scale \cite{prkachin2008structure}. VAS, SEN, AFF, OPI, and PSPI scales have ranges of $[0, 10], [0, 10], [0, 10], [0, 5], [0, 16]$, respectively, where $0$ indicates no pain, and $10$, $5$, and $16$ indicates maximum pain. 

Patients experienced different types of pain (e.g., bursitis, tendonitis), and $50\%$ of the patients took pain medications. To elicit painful expressions, patients performed eight active and passive range-of-motion tests (including abduction, flexion, and rotation) to affected and unaffected limbs in two different sessions. Similar to other affective (healthcare) datasets \cite{zhang2016multimodal}, this dataset is also highly imbalanced ($83.6\%$ of the frames do not represent pain expression, according to PSPI scale). Note that this study excluded patient $101$ from the experiments since the patient did not have samples with pain expression. 

\section{Experiments and Results}
\label{exprNRes}

\begin{figure}
\centering
\includegraphics[width=1\linewidth]{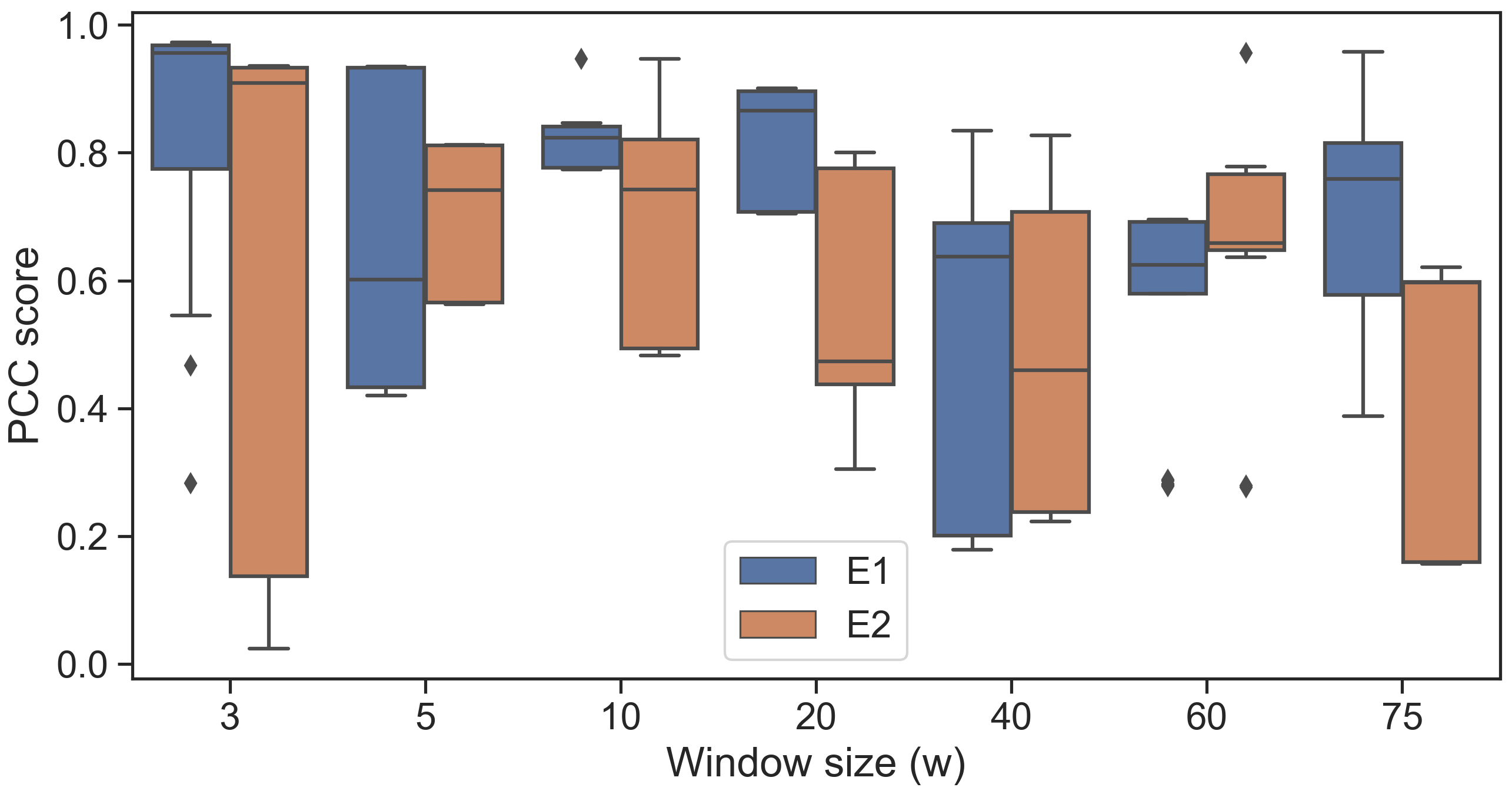}
\label{interpretModelTP}
\vspace{-4mm}
\caption{Statistical summary of correlation between TED and PSPI scores observed in experiments E1 and E2. The spread in the data is based on the participants in the studied dataset.}
\label{tedVsPSPI}
\vspace{-4mm}
\end{figure}

\textbf{Evaluation}. To quantitatively evaluate the effectiveness of the proposed TED algorithm, Pearson correlation coefficient (PCC), and significant test (p-value) were used. Qualitative analysis of TED and affect summarization (using TED) was demonstrated by visualizing subjective reports against the TED score. Finally, in the expectation and model interpretation case, to validate the pain classification model, leave-1-subject-out validation was performed and result was reported using F1-score.

\textbf{Benchmark.} It is important to note that previous works on facial dynamics focused on modeling multiple states of individual AUs and expressiveness works focused on video level expressiveness. In contrast, TED focuses on measuring expressiveness of a given video frame, not individual AUs. Considering this, direct comparison with previous works is not feasible with our experimental design. To the best of our knowledge, this is the first attempt towards quantifying temporal-expressiveness at video frame level, resulting in a baseline, on the UNBC-McMaster shoulder pain dataset \cite{lucey2011painful}.

\textbf{Multimodal Facial Features Tracking}. OpenFace \cite{baltruvsaitis2016openface}, a publicly available facial behavior analysis tool, was used to track facial features. This work tracked $L, H_o, H_r, G_l, G_r,$ and $I$. Since AUs are the core component of TED, we conducted two sets of experiments: \textbf{E1}: compute TED score using manually coded AUs along with openFace tracked $L, H_o, H_r, G_l$ and $G_r$; \textbf{E2}: compute TED score using predicted (via machine learning model) AUs along with openFace tracked $L, H_o, H_r, G_l$ and $G_r$. Since this work intended to study facial pain expression, this work considered pain expression related AUs only that are $4, 6, 9, 10, 25, 43$ \cite{werner2019automatic}. Note that openFace does not predict AU $43$; hence, in experiment E2, all pain-related AUs were used except AU $43$. 

\subsection{Quantitative Evaluation of TED} An ablation study was performed on window length $w$ to verify how far back temporal changes (i.e., dynamic information) need to be tracked. The following set was used, $w = \{3, 5, 10, 20, 40, 60, 75\}$, where $w=10$ is moving average $M$ over $10$ consecutive frames ($0.5$ second of data). Recall that PCC and p-value were used as evaluation metrics, to quantitively measure the similarity between the TED score and PSPI score. The reasoning behind comparing TED against PSPI is that, for a given frame, PSPI represents overall pain and the TED score represents overall pain expressiveness.

\begin{figure*}
\centering     
\subfigure[Experiment E1]{\label{fig:a}\includegraphics[width=.245\linewidth]{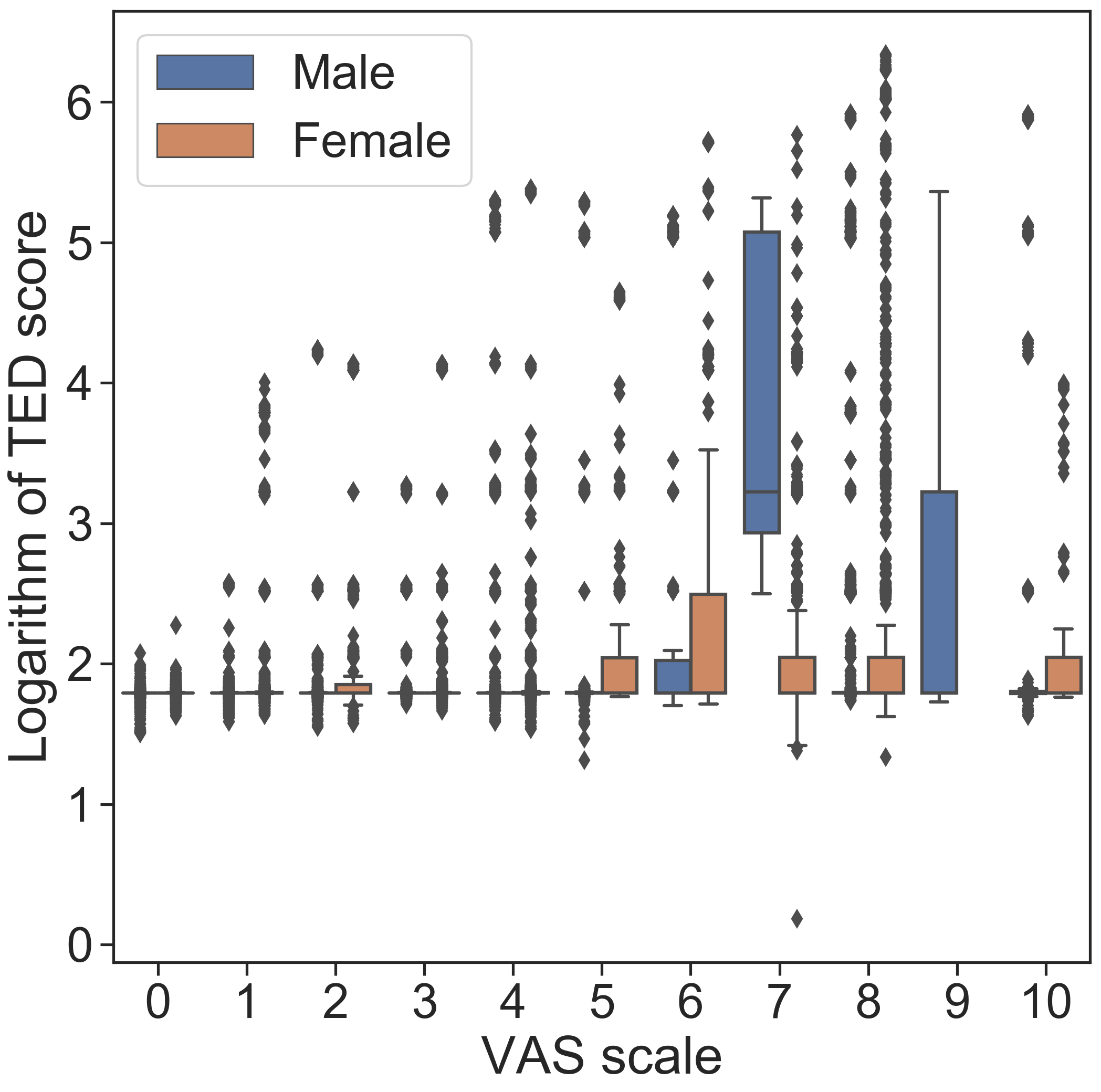}}
\label{as1}
\subfigure[Experiment E2]{\label{fig:b}\includegraphics[width=.245\linewidth]{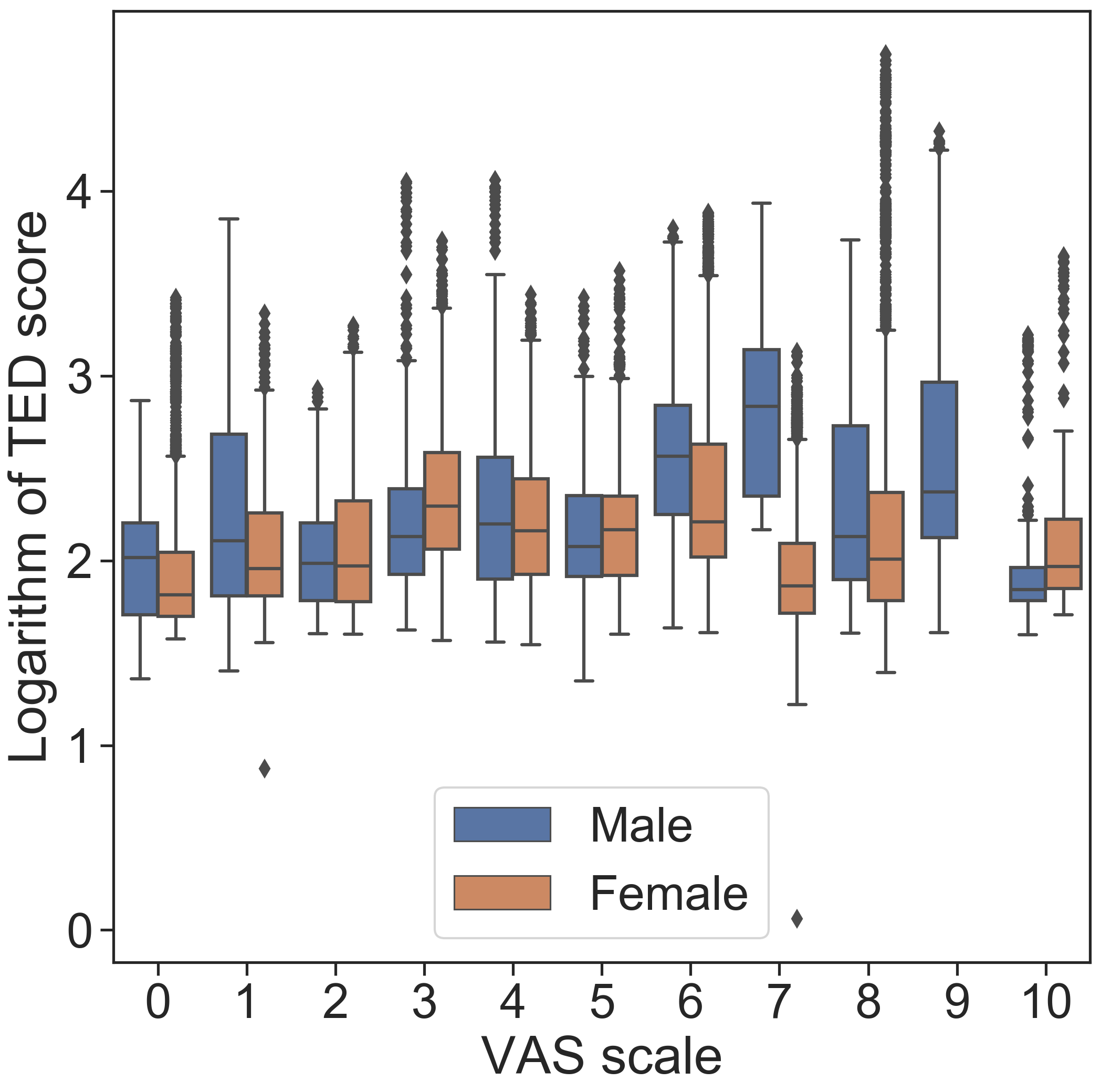}}
\label{as2}
\subfigure[Experiment E1]{\label{fig:a}\includegraphics[width=.245\linewidth]{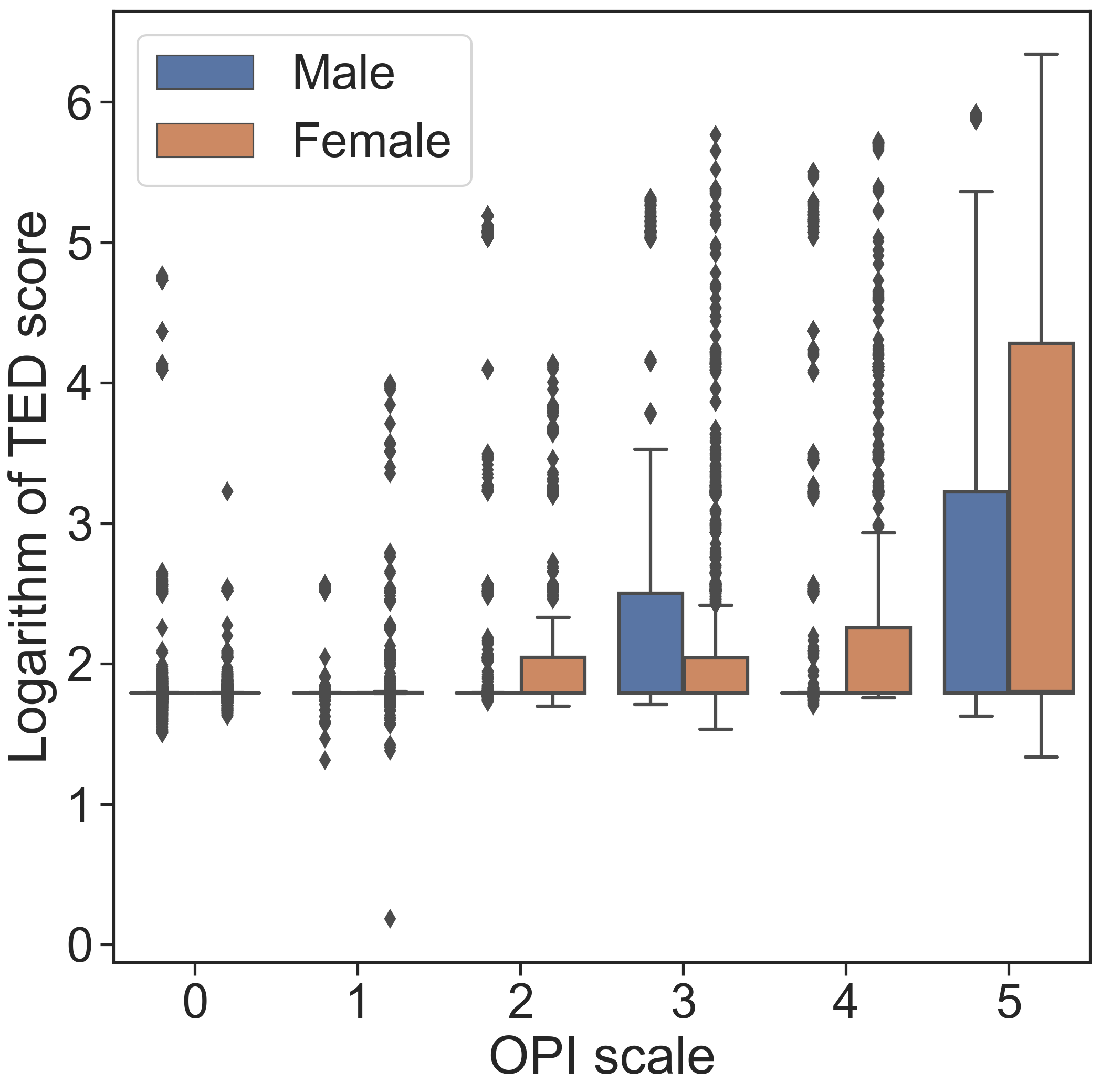}}
\label{as3}
\subfigure[FExperiment E2]{\label{fig:b}\includegraphics[width=.245\linewidth]{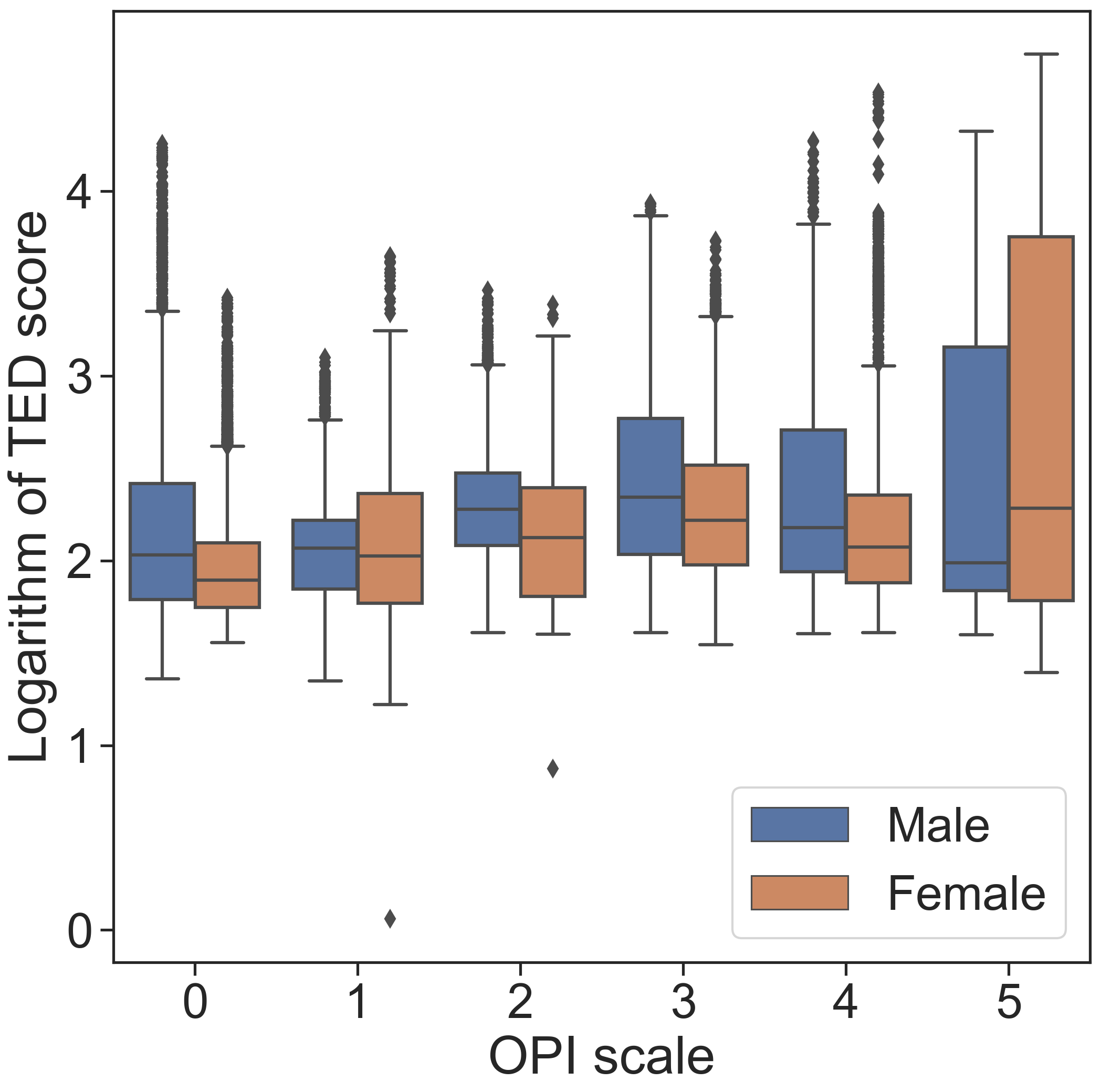}}
\label{as4}
\caption{Descriptive affect summary. Visualization of subjective pain scores against the logarithm of TED. To increase the readability of figure, logarithm of TED score was used instead of TED score itself. See Sec. \ref{exprNRes}(B) for details about the figure. }
\label{affSumm}
\vspace{-5mm}
\end{figure*}

The PCC and p-value were computed for each subject separately. From Fig. \ref{tedVsPSPI}, we infer that there is a positive correlation between the TED score and PSPI score for most participants. That is a strong indication that using the combination of static and dynamic facial information in TED can quantify the pain-specific expressiveness. We can also observe that there is a significant influence of $w$ on the performance of the algorithm; more precisely, using a $w$ that is too small or too large may not be useful due to the dynamic and fragile nature of the facial expressions. Based on the results shown in Fig. \ref{tedVsPSPI}, we can imply that to capture the dynamic component of expressiveness for a moment of time, more emphasis should be put on most recent ($[0.5, 1]$ second) facial information. In terms of PCC, it is noticeable from the median lines in Fig. \ref{tedVsPSPI} that in both E1 and E2, TED was able to compute the expressiveness with reasonable PCC values. Temporal-expressiveness was captured more accurately in experiment E1 compared to experiment E2. It is evident in Figs. \ref{as2} and \ref{as4} that TED failed to properly quantify the highly expressive frames in E2. The performance degradation in E2 could be explained, in part, by the error from the AU prediction models \cite{baltruvsaitis2016openface} as AU prediction models are not as good as expert (human) AU coders yet \cite{benitez2017emotionet}. For each of the experiments shown in Fig. \ref{tedVsPSPI}, the computed p-value was $\leq$ 0.005, except for participant $49$ (p-value $=0.21$ when $w=3$). In our manual investigation into video data, we observed that participant $49$ wore glasses which introduced occlusion that can degrade the AU prediction performance making the quantification of temporal-expressiveness difficult. 

To provide a formal baseline, we included the mean PCC value obtained across participants. The optimal PCC value was obtained when $w = 10$; in experiment E1 and E2, mean PCC (and p-value) scores were $0.75 (0)$ and $0.57 (0)$, respectively. The spread of the PCC values in Fig. \ref{tedVsPSPI} highlights the variability of pain-related expressiveness among participants. From that, we can conclude that individual differences among people in terms of pain elicitation could be a crucial factor in pain assessment based on visual data. Hence, in our next experiment, we focused on qualitative evaluation of TED by visualizing the relationship between TED and subjective reports such as self-reported and observer-reported pain scores in which we also incorporated gender as demographic context to evaluate gender-based variability.

\subsection{Qualitative Evaluation of TED and Affect Summarization} 
To provide a qualitative evaluation of TED and how TED could be used to mine insightful information, this work explored the relationship between subjective pain reports (VAS and OPI scales) and TED. For a given pain reporting scale, for each unit/category of the scale, the descriptive summary of the TED score was computed and visualized (Fig. \ref{affSumm}). Notice in Fig. \ref{affSumm} that the gender information was also incorporated as we are interested in depicting gender-specific variability in facial expressiveness \cite{hess2000emotional}. 

As can be seen in Figs. \ref{as1} and \ref{as3}, when the self and observer reports were $0$, the TED score of most frames across all subjects was approximately $6$ ($e^{1.8} \approx 6$). Hence, we can imply that in the context of pain expression, frames with TED score $6$ represents no pain expression. Another interesting finding is that no pain sequences also contain nominal frames with moderate to high TED score indicating that no pain samples contain pain frames as well. When the score is greater than $6$, the frames usually contain pain expression. Notice that when the VAS score and OPI score were in the ranges $[1, 5)$ and $[1, 3)$ (i.e., low to moderate pain), most sequences contain a small percentage of pain frames (TED score $> 6$). This suggests that the pain frames are outliers in those sequences. This could make classifying pain sequences from no pain sequences challenging due to the similarity issue \cite{werner2017analysis}. In contrast to low to moderate pain sequences, high intensity pain sequences contain a wide range of moderate to highly expressive frames, although a large number of frames are in the range of no pain to low pain. This suggests that the dataset is imbalanced in terms of expressiveness. This finding aligned with findings reported by Lucey et al. \cite{lucey2010automatically} in which the authors reported that only $16.3\%$ of frames in the dataset represent some form of pain expression.

This work also investigated whether gender plays any significant role in facial expressiveness. As we can see in Fig. \ref{affSumm}, the distribution of expressiveness (TED scores distribution) is quite different between males and females. For instance, female participants elicited facial pain expression for longer periods of time compared to male participants. However, a concluding remark of women are more expressive than men is not realistic given that the studied dataset contains only $25$ participants combined. These results are interesting and further investigation into this is necessary to gain more insight into the expressiveness of males versus females.

This work also explored the variability in reporters perspective using TED, and VAS and OPI scales. From Figs. \ref{as1} and \ref{as2}, it can be observed that the self-reported high intensity pain (VAS $ = 10$) sequences are less expressive than sequences with relatively less intensive pain (VAS score in between $7$ to $9$). This observation could be explained by the fact that in self-report, participants reported what they felt. Also keep in mind that the sample size of highest VAS scored sequences is relatively small which could skew the observation. On the other hand, in observer reported pain (OPI scale), it can be seen from Figs. \ref{as2} and \ref{as4} that when the observer observed high intensity pain expression (i.e. OPI score is high), TED based facial expressiveness score tends to be high as well. This phenomenon is understandable as observers usually report what they see (i.e., put more emphasis on the visual cues). Hence, our observations suggest that relying on visual information alone to predict affect states could be misleading to some extent. One possible solution would be combining the visual information with reporters' perspective, demographics, other modalities (e.g., physiology), and context to effectively model and predict affect.

\begin{figure*}
\centering     
\includegraphics[width=.245\linewidth]{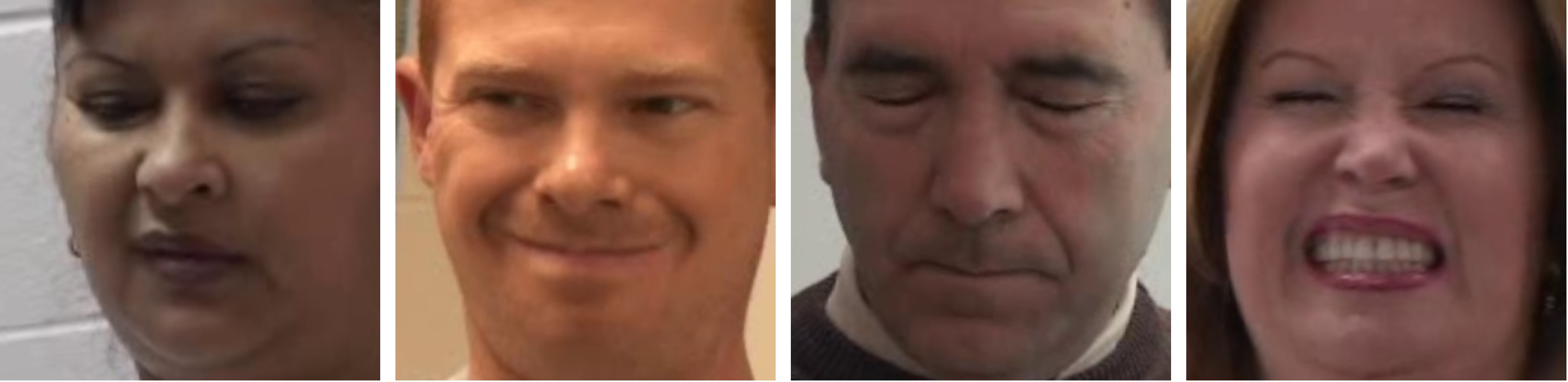}
\label{as1}
\includegraphics[width=.245\linewidth]{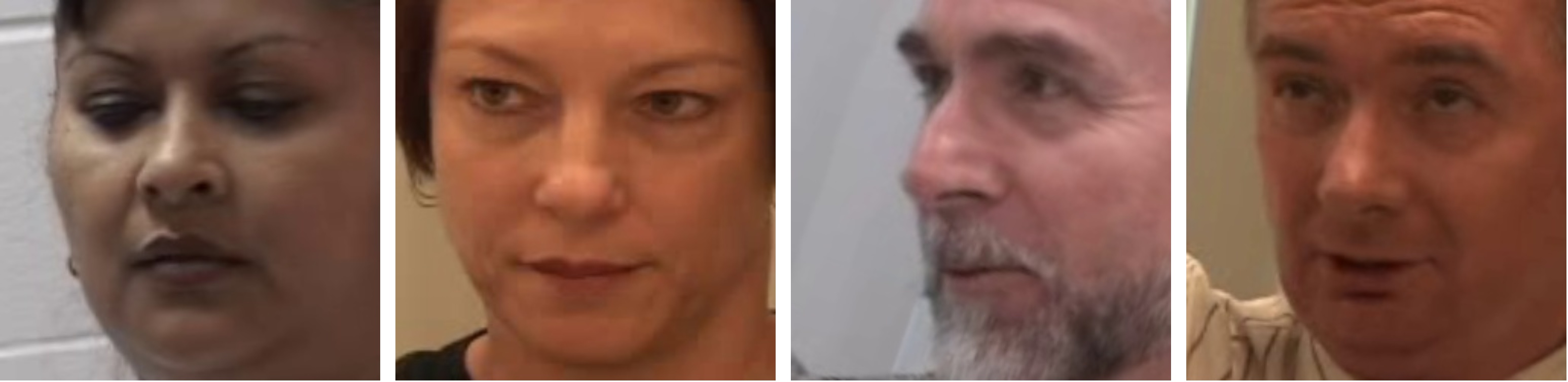}
\label{as2}
\includegraphics[width=.245\linewidth]{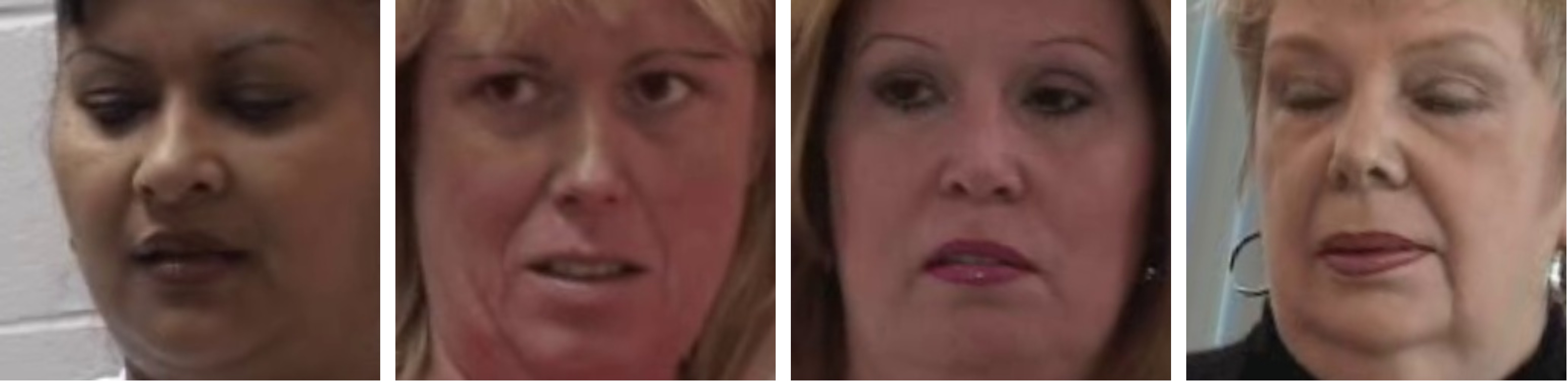}
\label{as3}
\includegraphics[width=.245\linewidth]{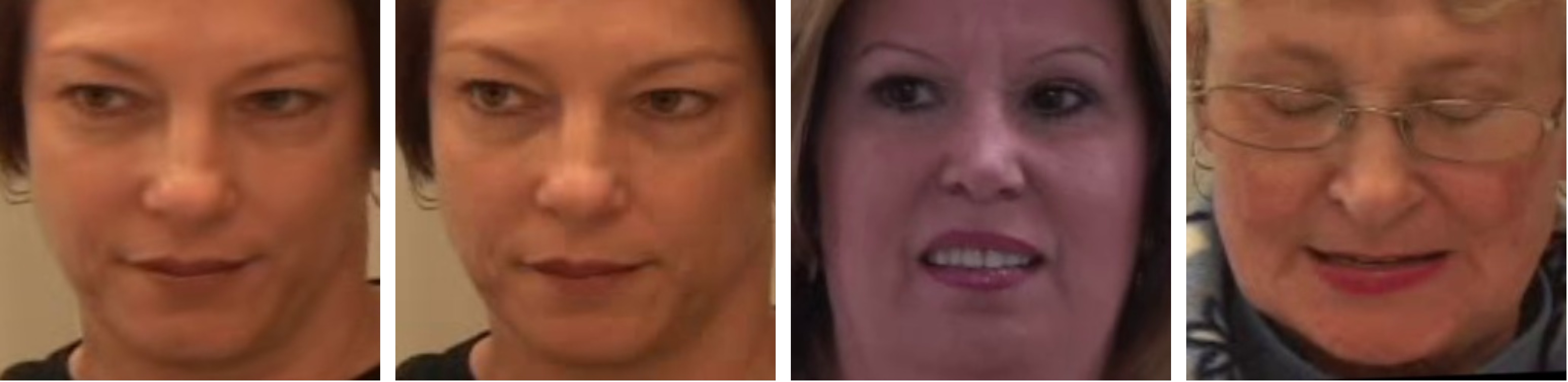}
\label{as4}
\subfigure[Scenario 1]{\label{fig:a}\includegraphics[width=.245\linewidth]{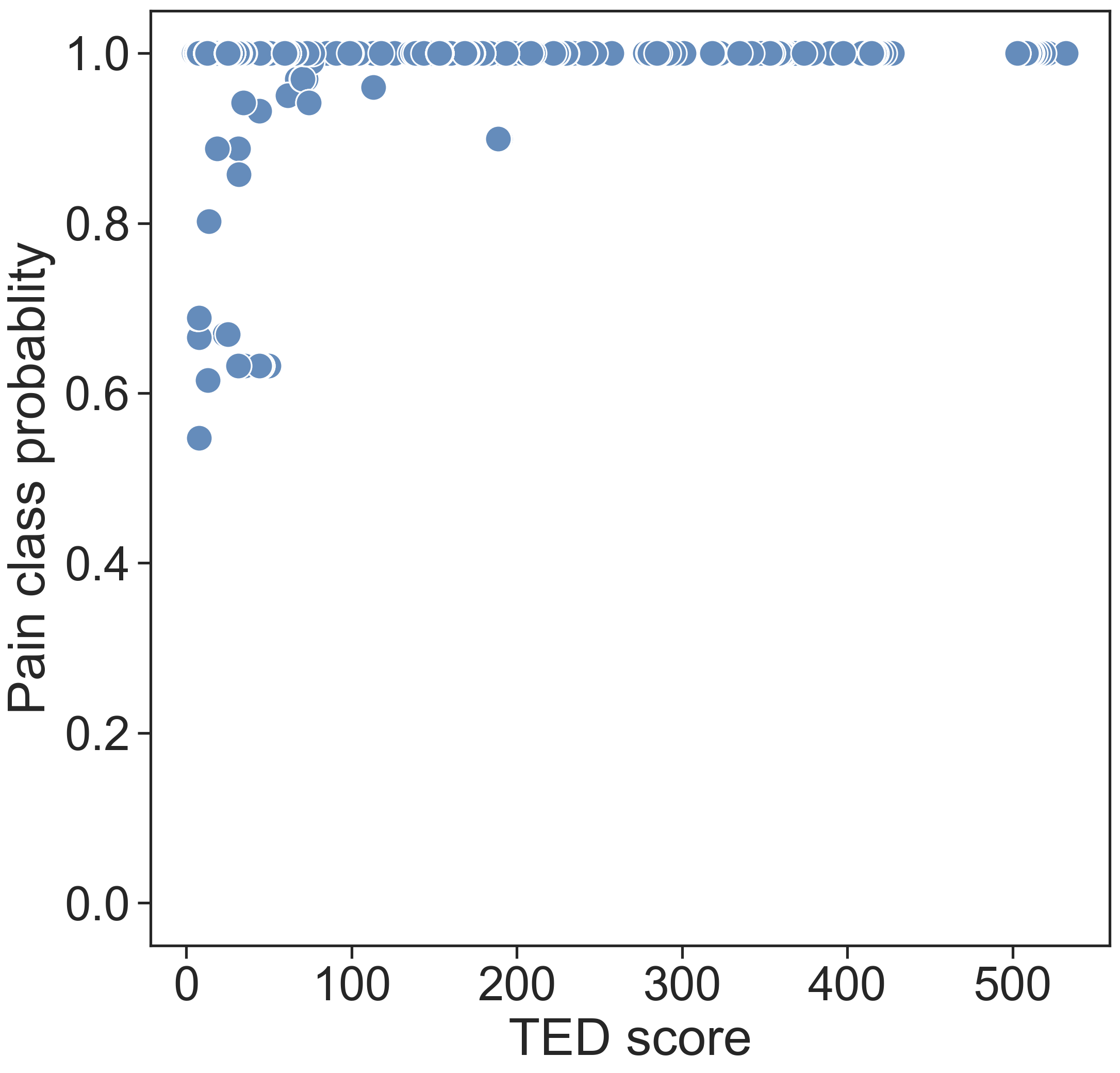}}
\label{interpretModelTP}
\subfigure[Scenario 2]{\label{fig:b}\includegraphics[width=.245\linewidth]{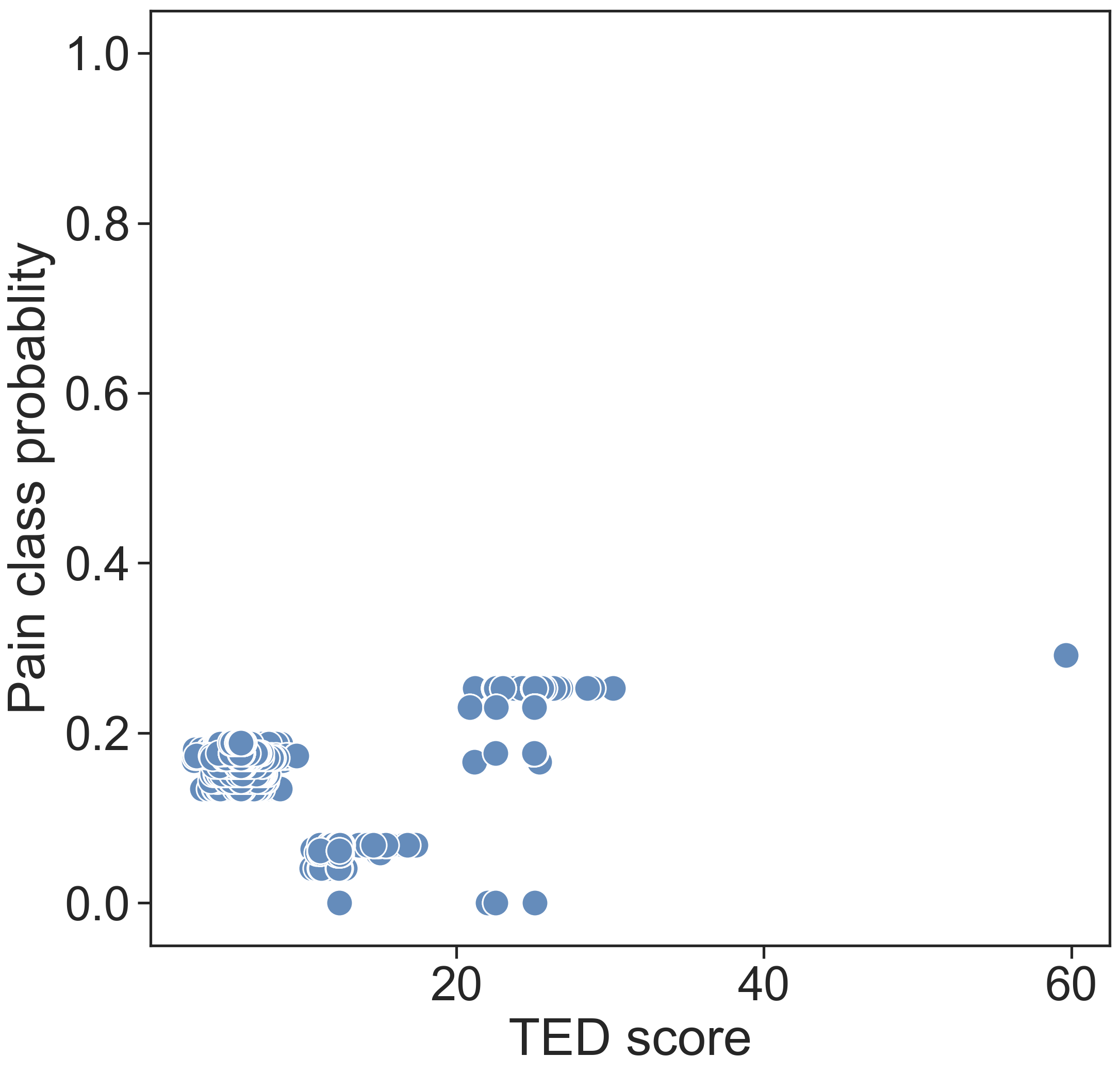}}
\label{interpretModelTN}
\subfigure[Scenario 3]{\label{fig:a}\includegraphics[width=.245\linewidth]{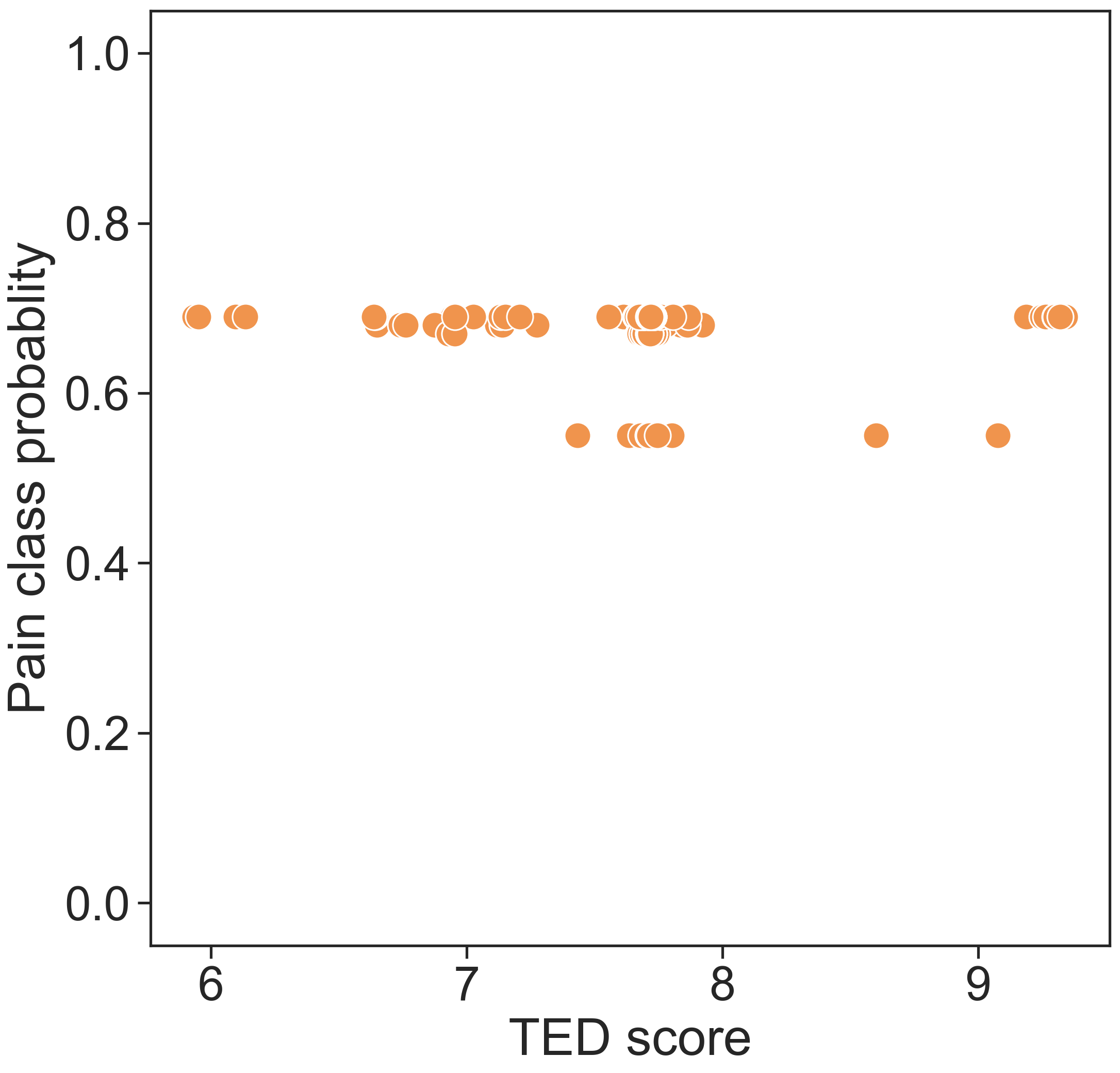}}
\label{interpretModelT1}
\subfigure[Scenario 4]{\label{fig:b}\includegraphics[width=.245\linewidth]{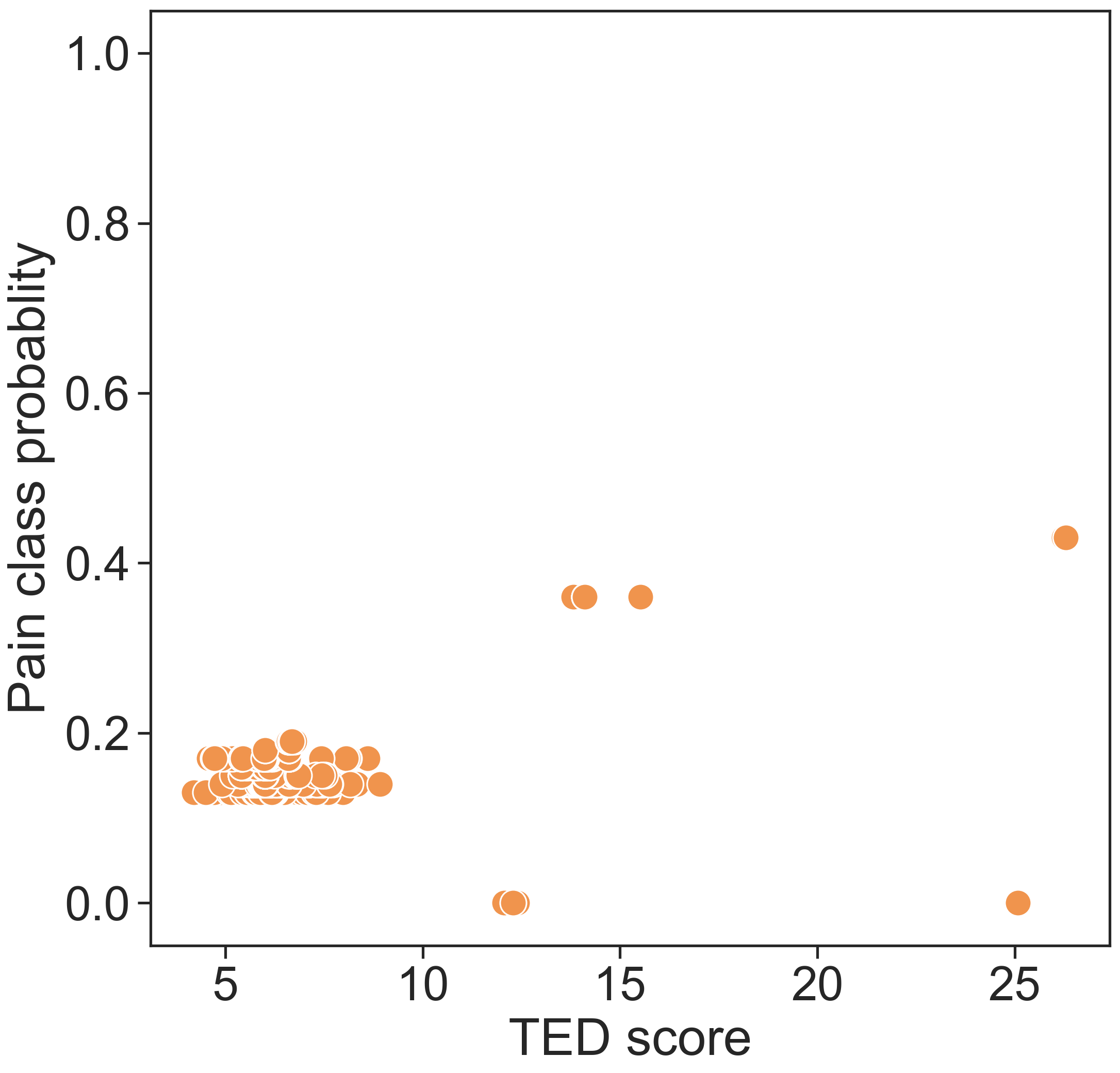}}
\label{interpretModelT2}
\caption{Model interpretation. The TED score vs. the confidence of pain classification model. Top row shows faces that correspond to the scenario in the plot below (e.g., the 4 faces above (c), are neutral faces that were classified as pain - type 1 error).}
\label{interpretModel}
\vspace{-5mm}
\end{figure*}

\subsection{Expectation and Model Interpretation} 

To provide experimental results of expectation and model interpretation, using TED, random forests algorithm \cite{breiman2001random} was trained to classify pain frames from no pain (neutral) frames using manually coded pain-related AU intensities. As mentioned earlier, leave-one-subject-out validation was used to validate the classifier and the pain classification result was reported using F1-score. The classifier obtained a mean F1-score of $0.86$ across participants. Note that to set up what we could expect from a trained pain classifier and to interpret the results produced by the classifier, this work only considered the confidence (probabilities) of positive class (pain), generated by the classifier. We considered four different scenarios: 1) \textit{scenario 1}: true positive - pain is classified as pain, 2) \textit{scenario 2}: true negative - neutral is classified as neutral, 3) \textit{scenario 3}: type 1 error - neutral is classified as pain, and 4) \textit{scenario 4}: type 2 error - pain is classified as neutral.

As it can be seen in Fig. \ref{interpretModelTP}, the classifier produced high confidence scores for frames with high TED scores, which was expected as in our case study, high TED score indicates facial pain expression. More concisely, for frames with TED scores at least $100$, the classifier was $100\%$ confident that frames represent pain expression. It can also be observed in Fig. \ref{interpretModelTP} that the faces visually show a lot of pain expression, resulting in high TED scores. Contrary to scenario 1, in scenario 2, the classifier was expected to produce lower probability scores (confidence) as samples belong to the neutral (negative) class (probability of negative class $= 1 -$ probability of positive class). In Fig. \ref{interpretModelTN}, it can be observed that when TED scores for the frames were low (e.g., $\leq 10$), the confidence of the classifier was also low (i.e., the classifier inferred with high confidence that frames belonged to neutral class). 

We also investigated the scenario in which the classifier makes the incorrect decision. For instance, in scenario 3, the model is expected to produce low probability for pain since frames belong to neutral. As it can be seen in Figs. \ref{as1} and \ref{as2}, neutral frames are likely to have TED scores of around $6$ ($e^{1.8} \approx 6$). However, it can be seen in Fig. \ref{interpretModelT1} that most neutral frames misclassified as pain frames have a TED score of at least $7$. The interpretation is that the classifier confused those neutral frames with very low-intensity pain frames. Finally, in scenario 4, even though frames belong to pain, the expectation from the classifier, in terms of confidence, is low since TED scores for most of these frames are low (i.e., frames represent pain with low expressiveness). As per the expectation, the classifier produced low pain class probabilities for most of these frames and, yet, ended up with incorrect prediction.

The incorrect prediction of pain can be summarized using TED based interpretation as follows: the classifier could get confused when expressiveness score for a given pain frame is close (i.e., low-intensity pain) to a neutral frame; the confusion was observed in scenarios 3 and 4. Further evidence can be observed in Fig. \ref{affSumm} - for both VAS and OPI scales, the facial expressiveness (TED scores) of low intensity pain samples was similar to the TED scores of no pain (neutral) samples. One interesting observation is that for one sample in Fig. \ref{interpretModelT2}, the disagreement between the TED score ($25)$ and classifier confidence ($0\%$) is high, which demands further investigation into TED, the classifier, and that specific sample. However, for the majority of samples, we observed a strong agreement between TED scores and classifier (see Fig. \ref{interpretModel}). It could be the case that those frames were wrongly annotated by FACS coder, which demands further investigation into the data. As you can see TED based interpretation of classifiers could lead us to build trustworthy and interpretable affect prediction models, which is useful as TED contains domain knowledge from psychology and affective computing \cite{ekman1978Facs, kring2007facial}.

\section{Discussion, Limitation, and Future Direction}
\label{DisLiF}

TED is a step towards alleviating the lack of quantified emotional expression modeling in affective computing. Our case study on spontaneous pain suggests that TED is capable of estimating facial expressiveness, and its utilities have the potential to improve the productivity of computing and application domain experts. Using TED, domain experts could inspect the relevant segments of data, which could boost their productivity. For instance, in Fig. 2(a), in the context of pain, the relevant frames (moment of interest) are in between $60$ to $120$. Experts, such as a doctor, can select the subset of frames in between $[80, 120]$ such as $80, 100, 120$, and inspect them manually to assess a patient instead of inspecting the entire sequence to speed up the inspection process. Recall that the applications of TED are not limited to affect summarization and model interpretation; as long as the video sequence contains a human face and the context of the problem at hand is aligned with affective computing (e.g., annotating sparse temporal data), TED could be relevant and useful.

Affect summarization based on TED could augment the rapid prototyping of an affective machine learning system, and ensure quick exploration of large amounts of data. This is essential as, in the era of big data, it is unrealistic to explore data manually. This work demonstrated the capability of TED to explore affect data along with gender, self-report, and observer reports to extract insights from unstructured data. This idea could be generalized to other demographic information (e.g., race, culture, age), reports (e.g., expert's opinion), and context (e.g., medical procedure). For instance, TED can be used to answer questions like for a given task $T$ in context $C$, how different, in terms of expressiveness, ethnic population $P_1$ aged in between $[O_1, O_2]$ is from ethnic population $P_2$ aged in between $[O_1, O_2]$, and how is it associated to different forms of affect reports? That would be useful to answer affective data science questions more efficiently.  

This work demonstrated how TED could be used to set up expectation and to interpret predictive results which has a major advantage, in the context of affective computing, over existing explainable AI methods as TED incorporates domain knowledge from psychology and affective computing \cite{ekman1978Facs, kring2007facial}. Even though experiments in this work were limited to pain modeling, it is extendable to other affective modeling tasks such as stress modeling \cite{uddin2019synthesizing}. In our future work, we intend to explore even more challenging model interpretation settings to test out the effectiveness of our approach given the necessity of explainable AI in sensitive application domains like healthcare. 

Predicted AU intensities are generally less accurate compared to manually coded AU intensities \cite{benitez2017emotionet}, which results in a limitation of TED as it depends on the quality of AU intensity prediction models in the absence of manually coded AUs. Although predicting AUs with high accuracy is still an open challenge \cite{benitez2017emotionet}, TED is still able to measure the expressiveness with minimal degradation. Another limitation is the lack of large affect datasets with manual coded AUs; most datasets contain a relatively small subset of manually coded AUs \cite{zhang2016multimodal}. Considering this, we will collect a new, large-scale dataset with manually coded AU intensities for each frame. We will explore information theory, dynamic time warping, and optical flow to compare against AU-based expressiveness. We will also explore how TED can be used to measure the misalignment between facial response and emotion.
\balance
\bibliographystyle{IEEEtran}
\bibliography{IEEEexample}

\begin{thebibliography}{10}
\providecommand{\url}[1]{#1}
\csname url@samestyle\endcsname
\providecommand{\newblock}{\relax}
\providecommand{\bibinfo}[2]{#2}
\providecommand{\BIBentrySTDinterwordspacing}{\spaceskip=0pt\relax}
\providecommand{\BIBentryALTinterwordstretchfactor}{4}
\providecommand{\BIBentryALTinterwordspacing}{\spaceskip=\fontdimen2\font plus
\BIBentryALTinterwordstretchfactor\fontdimen3\font minus
  \fontdimen4\font\relax}
\providecommand{\BIBforeignlanguage}[2]{{%
\expandafter\ifx\csname l@#1\endcsname\relax
\typeout{** WARNING: IEEEtran.bst: No hyphenation pattern has been}%
\typeout{** loaded for the language `#1'. Using the pattern for}%
\typeout{** the default language instead.}%
\else
\language=\csname l@#1\endcsname
\fi
#2}}
\providecommand{\BIBdecl}{\relax}
\BIBdecl

\bibitem{kring2007facial}
A.~M. Kring \emph{et~al.}, ``The facial exp coding system (faces): Dev,
  validation, and utility.'' \emph{Psych assessment}, vol.~19, no.~2, p. 210,
  2007.

\bibitem{gunes2011emotion}
H.~Gunes \emph{et~al.}, ``Emotion representation, analysis and synthesis in
  continuous space: A survey,'' in \emph{FG}, 2011, pp. 827--834.

\bibitem{zhang2016multimodal}
Z.~Zheng \emph{et~al.}, ``Multimodal spontaneous emotion corpus for human
  behavior analysis,'' in \emph{CVPR}, 2016, pp. 3438--3446.

\bibitem{cheng20184dfab}
S.~Cheng \emph{et~al.}, ``4dfab: A large scale 4d database for facial
  expression analysis and biometric applications,'' in \emph{CVPR}, 2018, pp.
  5117--5126.

\bibitem{walter2013biovid}
S.~Walter \emph{et~al.}, ``The biovid heat pain db data for the advancement and
  sys val of an auto pain rec sys,'' in \emph{CYBCO}, 2013, pp. 128--131.

\bibitem{walter2014automatic}
S.~Walter, S.~Gruss \emph{et~al.}, ``Automatic pain quantification using
  autonomic parameters,'' \emph{Psychology \& Neuroscience}, vol.~7, no.~3,
  2014.

\bibitem{williams2000simple}
A.~Williams \emph{et~al.}, ``Simple pain rating scales hide complex
  idiosyncratic meanings,'' \emph{Pain}, vol.~85, no.~3, pp. 457--463, 2000.

\bibitem{uddin2020multimodal}
M.~Uddin and S.~Canavan, ``Multimodal multilevel fusion for sequential
  protective behavior detection and pain estimation,'' in \emph{2020 IEEE FG}.

\bibitem{williamson2005pain}
A.~Williamson \emph{et~al.}, ``Pain: a review of 3 commonly used pain rating
  scales,'' \emph{Journal of clinical nursing}, vol.~14, no.~7, pp. 798--804,
  2005.

\bibitem{macari2018emotional}
S.~Macari \emph{et~al.}, ``Emotional expressivity in toddlers with autism
  spectrum disorder,'' vol.~57, no.~11, pp. 828--836, 2018.

\bibitem{wang2008automated}
P.~Wang \emph{et~al.}, ``auto video-based facial exp analysis of neuro
  disorders,'' \emph{J of neuroscience methods}, vol. 168, no.~1, pp. 224--238,
  2008.

\bibitem{minkel2011emotional}
J.~Minkel \emph{et~al.}, ``Emotional expressiveness in sleep-deprived healthy
  adults,'' \emph{Behavioral sleep medicine}, vol.~9, no.~1, pp. 5--14, 2011.

\bibitem{lindsey2019frequency}
E.~W. Lindsey, ``Frequency and intensity of emotional expressiveness and
  preschool children’s peer competence,'' \emph{The Journal of genetic
  psychology}, vol. 180, no.~1, pp. 45--61, 2019.

\bibitem{liu2019role}
M.~Liu \emph{et~al.}, ``The role of the face itself in the face effect: Sens,
  exp, and anticipated feedback in ind comp,'' \emph{Frontiers in Psych},
  vol.~9, 2019.

\bibitem{ekman1978Facs}
P.~Ekman and W.~Friesen, ``The facial action coding system: A tech. for the
  meas. of facial movement,'' \emph{Consulting Psychologists Press}, 1978.

\bibitem{lucey2011painful}
P.~Lucey \emph{et~al.}, ``Painful data: The unbc-mcmaster shoulder pain
  expression archive database,'' in \emph{FG}, 2011, pp. 57--64.

\bibitem{Lei2020}
S.~Lei \emph{et~al.}, ``Emotion or expressivity? an automated analysis of
  nonverbal perception in a social dilemma,'' in \emph{FG}, 2020.

\bibitem{jiang2013dynamic}
B.~Jiang \emph{et~al.}, ``A dynamic appearance descriptor approach to facial
  actions temporal modeling,'' \emph{IEEE trans on cyber}, vol.~44, 2013.

\bibitem{valstar2007combined}
M.~Valstar \emph{et~al.}, ``Combined svm and hmm for modeling facial action
  temporal dynamics,'' in \emph{Workshop on HCI}, 2007, pp. 118--127.

\bibitem{hammal2018facial}
Z.~Hammal \emph{et~al.}, ``Facial expressiveness in infants with and without
  craniofacial microsomia: preliminary findings,'' \emph{The Cleft
  Palate-Craniofacial Journal}, vol.~55, no.~5, pp. 711--720, 2018.

\bibitem{hammal2019dynamics}
Z.~Hammal, E.~Wallace \emph{et~al.}, ``Dynamics of face and head movement in
  infants with and without craniofacial microsomia: An auto approach,''
  \emph{Plastic and Reconstructive Surgery Global Open}, vol.~7, no.~1, 2019.

\bibitem{neubauer2017manual}
C.~Neubauer \emph{et~al.}, ``Manual and auto measures confirm—intranasal
  oxytocin increases facial expressivity,'' in \emph{ACII}, 2017.

\bibitem{guha2016computational}
T.~Guha \emph{et~al.}, ``A computational study of expressive facial dynamics in
  children with autism,'' \emph{IEEE Trans. on AC}, vol.~9, no.~1, 2016.

\bibitem{lin2019context}
V.~Lin \emph{et~al.}, ``Context-dependent models for predicting and
  characterizing facial expressiveness,'' \emph{arXiv preprint
  arXiv:1912.04523}, 2019.

\bibitem{werner2017analysis}
P.~Werner \emph{et~al.}, ``Analysis of facial expressiveness during
  experimentally induced heat pain,'' in \emph{ACIIW}, 2017, pp. 176--180.

\bibitem{faraj2020facially}
Z.~Faraj \emph{et~al.}, ``Facially express human robot face,''
  \emph{HardwareX}, 2020.

\bibitem{dahmani2019conditional}
S.~Dahmani \emph{et~al.}, ``Conditional variational auto-encoder for
  text-driven expressive audiovisual speech synthesis.'' in \emph{INTERSPEECH},
  2019.

\bibitem{zhang2014bp4d}
X.~Zhang \emph{et~al.}, ``Bp4d-spontaneous: a high-res spon 3d dynamic facial
  exp db,'' \emph{IVC}, vol.~32, no.~10, pp. 692--706, 2014.

\bibitem{valstar2015fera}
M.~Valstar \emph{et~al.}, ``Fera 2015-second facial expression recognition and
  analysis challenge,'' in \emph{FG}, vol.~6, 2015, pp. 1--8.

\bibitem{tang20083d}
H.~Tang and T.~S. Huang, ``3d facial exp rec based on properties of line
  segments connecting facial feature points,'' in \emph{FG}, 2008, pp. 1--6.

\bibitem{zhang2013high}
X.~Zhang \emph{et~al.}, ``A high-res spon 3d dyn facial exp db,'' in \emph{FG},
  2013.

\bibitem{mavadati2013disfa}
M.~Mavadati \emph{et~al.}, ``Disfa: A spontaneous facial action intensity
  database,'' \emph{IEEE Transactions on AC}, vol.~4, no.~2, pp. 151--160,
  2013.

\bibitem{mavadati2016extended}
M.~Mavadati, P.~Sanger \emph{et~al.}, ``Extended disfa dataset: Investigating
  posed and spontaneous facial expressions,'' in \emph{CVPRW}, 2016.

\bibitem{yan2019feafa}
Y.~Yan \emph{et~al.}, ``Feafa: A well-annotated dataset for facial expression
  analysis and 3d facial animation,'' in \emph{ICMEW}, 2019, pp. 96--101.

\bibitem{haque2018deep}
M.~Haque \emph{et~al.}, ``Deep mm pain rec: a database and comparison of
  spatio-temporal visual modalities,'' in \emph{FG}, 2018, pp. 250--257.

\bibitem{aung2015automatic}
M.~Aung \emph{et~al.}, ``The auto detect of chron pain-related exp: reqs, chals
  and the mm emopain dataset,'' \emph{IEEE Trans. on AC}, vol.~7, no.~4, 2015.

\bibitem{baltrusaitis2013constrained}
T.~Baltrusaitis \emph{et~al.}, ``Constrained local neural fields for robust
  facial landmark detection in the wild,'' in \emph{ICCVW}, 2013, pp. 354--361.

\bibitem{bassili1979emotion}
J.~Bassili, ``Emo rec: The role of fac move and the rel imp of up and low areas
  of the face.'' \emph{J. of per and soc psych}, vol.~37, no.~11, 1979.

\bibitem{knight1997role}
B.~Knight and A.~Johnston, ``The role of movement in face recognition,''
  \emph{Visual cognition}, vol.~4, no.~3, pp. 265--273, 1997.

\bibitem{baltruvsaitis2016openface}
T.~Baltru{\v{s}}aitis \emph{et~al.}, ``Openface: an open source facial behavior
  analysis toolkit,'' in \emph{WACV}, 2016, pp. 1--10.

\bibitem{awad2019role}
D.~Awad \emph{et~al.}, ``The role of emotional expression and eccentricity on
  gaze perception,'' \emph{Frontiers in psychology}, vol.~10, p. 1129, 2019.

\bibitem{wu2019continuous}
S.~Wu \emph{et~al.}, ``Continuous emotion recognition in videos by fusing
  facial expression, head pose and eye gaze,'' in \emph{ICMI}, 2019, pp.
  40--48.

\bibitem{baltruvsaitis2015cross}
T.~Baltru{\v{s}}aitis \emph{et~al.}, ``Cross-dataset learning and
  person-specific norm. for auto. au detection,'' in \emph{FG}, vol.~6, 2015,
  pp. 1--6.

\bibitem{hadid2011analyzing}
A.~Hadid, ``Analyzing fac behav feat from vids,'' in \emph{HBU}, 2011.

\bibitem{barrett2019emotional}
L.~Barrett \emph{et~al.}, ``Emo exp recon: challenges to infer emotion from
  human facial move,'' \emph{Psych Science in the Pub Int}, vol.~20, no.~1,
  2019.

\bibitem{krumhuber2013effects}
E.~Krumhuber \emph{et~al.}, ``Effects of dynamic aspects of facial expressions:
  A review,'' \emph{Emotion Review}, vol.~5, no.~1, pp. 41--46, 2013.

\bibitem{tan2018introduction}
P.~Tan, \emph{Introduction to data mining}.\hskip 1em plus 0.5em minus
  0.4em\relax Pearson Education India, 2018.

\bibitem{diez2012openintro}
D.~Diez \emph{et~al.}, \emph{OpenIntro statistics}.\hskip 1em plus 0.5em minus
  0.4em\relax OpenIntro, 2012.

\bibitem{prkachin2008structure}
K.~Prkachin \emph{et~al.}, ``The struc, rel and val of pain exp: Evidence from
  patients with shoulder pain,'' \emph{Pain}, vol. 139, no.~2, pp. 267--274,
  2008.

\bibitem{werner2019automatic}
P.~Werner \emph{et~al.}, ``Automatic recognition methods supporting pain
  assessment: A survey,'' \emph{IEEE Transactions on AC}, 2019.

\bibitem{benitez2017emotionet}
C.~Benitez-Quiroz \emph{et~al.}, ``Emotionet challenge: Rec. of facial
  expressions of emotion in the wild,'' \emph{arXiv preprint arXiv:1703.01210},
  2017.

\bibitem{hess2000emotional}
U.~Hess \emph{et~al.}, ``Emotional expressivity in men and women: Stereotypes
  and self-perceptions,'' \emph{Cog \& Emo}, vol.~14, no.~5, pp. 609--642,
  2000.

\bibitem{lucey2010automatically}
P.~Lucey \emph{et~al.}, ``Automatically detecting pain in video through facial
  action units,'' \emph{IEEE Trans. on SMC (Cybernetics)}, vol.~41, no.~3,
  2010.

\bibitem{breiman2001random}
L.~Breiman, ``Random forests,'' \emph{Machine learning}, 2001.

\bibitem{uddin2019synthesizing}
M.~T. Uddin and S.~Canavan, ``Synthesizing physiological and motion data for
  stress and meditation detection,'' in \emph{ACIIW}, 2019, pp. 244--247.

\end{thebibliography}


\end{document}